\pdfoutput=1 
\documentclass[sigconf]{aamas} 
\usepackage{balance} % for balancing columns on the final page

\setcopyright{none}
\acmConference[MODeM '21]{Proc.\@ of the 1st Multi-Objective Decision Making Workshop (MODeM 2021)}
{July 14-16, 2021}{Online, \url{http://modem2021.cs.nuigalway.ie}}{Hayes, Mannion, Vamplew (eds.)}
\copyrightyear{2021}
\acmYear{2021}
\acmDOI{}
\acmPrice{}
\acmISBN{}
\usepackage{alphabeta}
\usepackage{centernot}  % for the use of \centernot
\usepackage{xcolor}
\newcommand{\red}[1]{{\color{red} #1}}

\newcommand{\todo}[1]{{\color{red} #1}}

\usepackage{pgfplotstable}
\usepackage{pgfplots}   % used instead of tikz, to allow plotting
\pgfplotsset{compat=1.16}   % to use latex arrows without warning message
\usetikzlibrary{external}   % to cache figures
\usetikzlibrary{backgrounds}

\usepackage{tabularx,hhline}   % Automatic calculation of column widths
\usepackage{bigstrut}
\usepackage{tablefootnote}
\usepackage{multirow}
\usepackage{booktabs}% http://ctan.org/pkg/booktabs
\usepackage{makecell}% http://ctan.org/pkg/makecell
\usepackage[mathscr]{euscript}
\usepackage{setspace}  % for spacing in algorithm environment
\usepackage[vlined,ruled,linesnumbered]{algorithm2e}
\SetKwFor{RepTimes}{repeat}{times}{end}
\usepackage{cleveref}
\Crefname{figure}{Figure}{Figures}

\usepackage{graphicx, caption, subcaption}
\usepackage[export]{adjustbox}

\DeclareMathOperator*{\Argmax}{arg\,max}
\DeclareMathOperator*{\Argmin}{arg\,min}

\newcommand{\indicator}{\mathbf{1}}
\newcommand{\sep}{\,;}

\newcommand{\transfn}{\mathscr{T}}

\usepackage{stmaryrd}
\newcommand{\nrange}[2]{[#1..#2]}%{\llbracket#1..#2\rrbracket}%{\mathbb{\overline N}_{#1:#2}}

\usepackage{xspace}
\newcommand{\Vp}{P}
\newcommand{\Vr}{V}
\newcommand{\Qp}{\mathscr{P}}
\newcommand{\Qr}{Q}

\newcommand{\optQp}[1]{\mathscr{\hat P}_{#1}}
\newcommand{\optQr}[1]{{\hat Q}_{#1}}

\newcommand{\optA}[1]{\mathscr{\hat A}_{#1}}
\newcommand{\optAc}[1]{\mathscr{\hat A}_{#1}{\!\!\!\;\!\!}{'}\,}
\newcommand{\optAcinarg}[1]{\mathscr{\hat A}_{#1}{\!\!\!\;\!\!}^{'}\,}

\newcommand{\safeset}{\mathcal{S}}
\newcommand{\unsafeset}{\mathcal{F}}
\newcommand{\optsafeset}{\mathcal{\hat S}}
\newcommand{\optunsafeset}{\mathcal{\hat F}}
\newcommand{\optpolicy}{{\hat \pi}_{\theta}}
\newcommand{\optpolicyvar}{{\hat \pi}_{{\vartheta}}}
\newcommand{\optpolicyvarp}{{\hat \pi}_{{\vartheta'}}}

\newcommand{\LRstyle}[1]{\mathsf{#1}}
\newcommand{\Left}{\LRstyle{L}}
\newcommand{\Right}{\LRstyle{R}}
\newcommand{\LL}{\LRstyle{LL}}
\newcommand{\RR}{\LRstyle{RR}}
\newcommand{\LR}{\LRstyle{LR}}
\newcommand{\RL}{\LRstyle{RL}}

\acmSubmissionID{???}

\title[Recursive Constraint RL]{Recursive Constraints to Prevent Instability \\ in Constrained Reinforcement Learning}\thanks{*\,Contributed equally}

\author{Jaeyoung Lee*}
\affiliation{
  \institution{University of Waterloo}
  \city{Waterloo, ON, Canada}}
\email{jaeyoung.lee@uwaterloo.ca}

\author{Sean Sedwards*}
\affiliation{
  \institution{University of Waterloo}
  \city{Waterloo, ON, Canada}}
\email{sean.sedwards@uwaterloo.ca}

\author{Krzysztof Czarnecki}
\affiliation{
 \institution{University of Waterloo}
 \city{Waterloo, ON, Canada}}
\email{krzysztof.czarnecki@uwaterloo.ca}

\begin{abstract}
We consider the challenge of finding a deterministic policy for a Markov decision process that uniformly (in all states) maximizes one reward subject to a probabilistic constraint over a different reward.
Existing solutions do not fully address our precise problem definition, which nevertheless arises naturally in the context of safety-critical robotic systems.
This class of problem is known to be hard, but the combined requirements of determinism and uniform optimality can create learning instability.

In this work, after describing and motivating our problem with a simple example, we present a suitable constrained reinforcement learning algorithm that prevents learning instability, using recursive constraints.
Our proposed approach admits an approximative form that improves efficiency and is conservative w.r.t. the constraint.
\end{abstract}


\newcommand{\BibTeX}{\rm B\kern-.05em{\sc i\kern-.025em b}\kern-.08em\TeX}

\begin{document}

%%% The following commands remove the headers in your paper. For final 
%%% papers, these will be inserted during the pagination process.

\pagestyle{fancy}
\fancyhead{}

%%% The next command prints the information defined in the preamble.

\maketitle 

%%%%%%%%%%%%%%%%%%%%%%%%%%%%%%%%%%%%%%%%%%%%%%%%%%%%%%%%%%%%%%%%%%%%%%%%

\section{Introduction}
Constrained optimization of Markov decision processes is a well-studied field, with a number of algorithms in existence for specific problem definitions~\cite{Altman1999,Achiam2017,Tessler2018,Chow2018,Feinberg1999, Dolgov2005,Geibel2005,Feinberg2000}.
In our work, we wish to find a deterministic policy that uniformly maximizes one reward subject to a probabilistic constraint over a different reward.
This problem arises naturally in safety-critical systems, such as autonomous driving, where it is required to maximize performance while bounding the probability of hazards, in all states.
Since these systems are our focus, we will refer to the notion of satisfying a constraint as safety.
Hence, our optimality can be intuitively stated as: in every state, an agent should choose a safe action that maximizes performance or, if no safe action exists, the least unsafe action.
Despite its apparent simplicity and similarity to other problems, this definition appears not to have an adequate existing solution.

The characteristics that distinguish our problem specification from previous work is that we require both a deterministic policy (for explainability) and uniform optimality (optimality in every state).
Moreover, to ensure safety, the calculation of probability cannot be discounted, as it can with other rewards.
In what follows, we show that with naive reinforcement learning algorithms, this combination entails learning instability (oscillation) and inaccurate estimates of probability.

Much previous work has considered stochastic policies~\cite{Feinberg1999,Achiam2017,Tessler2018,Chow2018}, while that which considers deterministic policies~\cite{Feinberg2000,Geibel2005,Dolgov2005} does not include the notion of uniform optimality.
Unpublished preprints that appear to address a similar problem formulation to ours~\cite{Undurti2011,Kalweit2020} use examples that do not demonstrate the inherent instability that we have discovered.
This is perhaps understandable: even with an adequate example, we have observed that a suitable choice of hyperparameters with a naive learning approach may sufficiently mask the instability for it to seem like mere stochasticity. This is likely to be especially true when using function approximation, where the expectation of convergence is less.

Our proposed solution (\Cref{sec:recursive-constraints}) is a model-based reinforcement learning algorithm that uses recursive constraints to prevent instability while achieving our notion of optimality.
Our proposed approach admits an approximative form that improves efficiency and is conservative w.r.t. safety.
We suggest this algorithm as a suitable candidate for further extension to the model-free case.

The remainder of this paper proceeds as follows.
In~\Cref{sec:preliminaries} we define the mathematical preliminaries required for the sequel.
In~\Cref{sec:constrained-optimality} we describe the (sometimes conflicting) requirements of our notion of constrained optimality.
In~\Cref{sec:pathological-behaviours} we give a simple motivating example that demonstrates the inherent instability with naive learning approaches.
In~\Cref{sec:recursive-constraints} we describe our algorithm based on recursive constraints and provide results of experiments in~\Cref{sec:experiments} that demonstrate its advantages.
We briefly conclude our work in~\Cref{sec:conclusion}.

% A sequence $\langle a_k \rangle_{k \in \mathbb{N}_0}$ is abbreviated as $\langle a_k \rangle$ for simplicity.
%$2^X$ denotes the power set of $X$. Given a probability space $(X, 2^X, \Prob)$ and $H \subseteq X$ s.t. $\Prob(H) > 0$, $\Prob(Y | H) := \Prob( Y \cap H) / \Prob(H)$ denotes the conditional probability of $Y \subseteq X$ on $H$.

%We say that $X$ is partitioned by its subsets $X_1$, $X_2$, $\dots$, $X_n$ iff $X_i$ and $X_j$ are disjoint for all $i, j \in \{1,2,\dots, n\}$ and $X = \bigcup_{i=1}^n X_i$. 

\section{Preliminaries}
\label{sec:preliminaries}
%In this paper, we consider a finite Markov decision process (MDP) $\mathscr{M} := (\mathscr{ S}, \mathscr{A}, \mathscr{P}, \gamma, \mathscr{R}, \mathscr{R}_\perp)$ for a state space $\mathscr{ S} := \mathscr{S \cup S_\perp}$ consisting of the disjoint sets of all non-terminal states $\mathscr{S}$ and all terminal ones $\mathscr{S}_\perp$, an action space $\mathscr{A}$, and a transition function $\mathscr{P}: \mathscr{S \times A} \to \Dist(\mathscr{ S})$ describing the distribution of the next state $s' \in \mathscr{ S}$ given the current state $s \in \mathscr{S}$ and chosen action $a \in \mathscr{A}$. The spaces $\mathscr{ S}$ and $\mathscr{A}$ are finite. We denote $\mathscr{A}(s)$ for $s \in \mathscr{S}$ the set of all available actions at state~$s$, i.e., 

%In mathematical statements, iff stands for ``if~and~only~if'' and s.t. for ``such that.'' 
In this paper, $\mathbb{\overline N} := \mathbb{N} \cup \{ \infty \}$ and $\mathbb{\overline N}_0 := \mathbb{N}_0 \cup \{ \infty \}$ denote the sets of extended natural numbers and extended non-negative integers, respectively. For $n, m \in \mathbb{\overline N}_0$, we also denote 
\begin{align*}
	\nrange{n}{m} := \{ k \in \mathbb{\overline N}_0 \, \vert \, n \leq k \leq m \}
\end{align*}
%$\Min{n}{m} := \min(n, m)$ and $\Max{n}{m} := \max(n, m)$. 
%A sequence of mapping $\langle \pi_n \rangle$ between finite sets is said to converge to $\pi$ iff there exists $N \in \mathbb{N}$ s.t. $\pi_n = \pi$ for all $n \geq N$.
%A distribution on a finite set $X$ is a function $p: X \to [0, 1]$ s.t. $\sum_{x \in X} p(x) = 1$. $\Dist(X)$ denotes the set of all distributions on $X$. 

We consider a finite Markov decision process (MDP) 
\[
	\mathscr{M} := (\mathscr{S^+}, \mathscr{A^+}, \transfn, \gamma, \mathscr{R})
\] 
where $\mathscr{S^+} := \mathscr{S \cup S_\perp}$ is a finite set of states consisting of the disjoint sets of all non-terminal states $\mathscr{S}$ and all terminal states $\mathscr{S}_\perp$. $\mathscr{A^+}$ is a finite set of actions and $\gamma \in [0, 1)$ is discount rate; transition function $\transfn(s, a)$ describes the distribution of next state over $\mathscr{S^+}$, given a current state $s \in \mathscr{S}$ and a chosen action $a \in \mathscr{A^+}$; the reward model $\mathscr{R} : \mathscr{S}^+ \times \mathscr{A^+} \times \mathscr{S}^+ \to \mathbb{R}$ determines the reward $\mathscr{R}(s, a, s')$ for transition $(s, a, s') \in \mathscr{S} \times \mathscr{A^+} \times \mathscr{S}^+$ and the terminal one $\mathscr{R}(s, a, s)$ at terminal state-action $sa \in \mathscr{S}_\perp \times \mathscr{A}^+$. The sets $\mathscr{S}^+$ and $\mathscr{A}^+$ are finite. We denote $\mathscr{A}(s)$ ($\subseteq \mathscr{A}^+$) the set of all available actions in state~$s \in \mathscr{S}^+$, 
\iffalse
i.e., 
$
	\mathscr{A}(s) := \big \{ a \in \mathscr{A} \, \vert \, \transfn(s, a) \text{ is defined or } s \in \mathscr{S_\perp} \big \}
$, 
\fi
which is assumed non-empty for all $s \in \mathscr{S^+}$.
\iffalse
and simplified to $\mathscr{A}(s)= \mathscr{A}$ for all $s \in \mathscr{S}_\perp$.
\fi

A \emph{path} is a sequence of alternating states, actions and rewards
\[
	(s_0 a_0 r_0) (s_1 a_1 r_1) \cdots (s_{T - 1} a_{T-1} r_{T-1}) s_{T}a_{T} r_{T}
\]
s.t.
$
	s_t a_t \in \mathscr{S} \times \mathscr{A}(s_t)$, $s_{t+1} \!\sim\! \transfn(s_t, a_t) 
$ and $r_{t} = \mathscr{R}(s_t, a_t, s_{t+1})$ for each $t \in \nrange{0}{T \!-\! 1}$, $s_T a_T \in \mathscr{S}_\perp \times \mathscr{A}(s_T)$ and $r_T = \mathscr{R}(s_T, a_T, s_T)$, where $T \in \mathbb{\overline N}_0$ denotes the terminal index, the first hitting time on $\mathscr{S}_\perp$. The reward sequence $r_0r_1\cdots r_T$ and discount rate $\gamma$ define the return
\begin{equation}
	R_T := r_0 + \gamma \cdot r_1 + \gamma^2 \cdot r_2 + \gamma^3 \cdot r_3 + \cdots + \gamma^{T} \cdot r_{T}
	\label{eq:return}
\end{equation}

A \emph{policy} is a mapping ${\pi: \mathscr{S^+} \to \mathscr{A}^+}$ s.t. $\pi(s) \in \mathscr{A}(s)$ for all $s \in \mathscr{S^+}$. 
%$\Pi$ denotes the set of all policies. 
%A policy~$\pi$ is said to be \emph{deterministic} iff $\pi(s)(a)\in\{0,1\}$ for all $(s, a) \in \mathscr{S} \times \mathscr{A}$. 
Given $s \in \mathscr{S^+}$ (resp. $sa \in \mathscr{S^+} \times \mathscr{A}(s)$), policy~$\pi$ and MDP $\mathscr{M}$ generate paths s.t. $s_0 = s$ (resp. $s_0a_0 = sa$) and $a_t = \pi(s_t)$ thereafter, thus inducing probability measures over all such paths. 
%In a similar manner, given $s$ (resp. $sa$), stationary policy~$\pi$ and $\mathscr{M}$ generate infinite-horizon paths $h_\infty$ s.t. for all $n \in [T]$, $s_0 = s$ and $a_n = \pi(s_n)$ (resp. $s_0a_0 = sa$ and ${a_{n + 1} = \pi(s_{n + 1})}$), inducing respective probability measures over all such paths. 
For notational simplicity, we adopt the notations 
\[
	\mathbb{P} ( \varphi \, | \, s_0 = s, \, \pi ) \text{ and } \mathbb{P} ( \varphi \, | \, s_0 a_0 = s a,\, \pi ),
\]
denoting the probabilities that the paths generated by policy $\pi$, given $s_0 = s$ and ${s_0a_0 = sa}$, respectively, satisfy the property $\varphi$; 
\[
	\mathbb{E} ( x \, | \, s_0 = s, \, \pi ) \text{ and } \mathbb{E} ( x \, | \, s_0 a_0 = s a, \, \pi )
\]
denote the corresponding expectations of a random variable~$x$. 
%We also denote the set of all policies by $\Pi$  and all deterministic ones by $\Pi_\mathrm{det}$ ($\subset \Pi$).

\paragraph{Probabilistic Reachability of Failure States}
Let $\mathscr{F}_\perp \subseteq \mathscr{S}_\perp$ be a set of all failure states, then given policy $\pi$ defines the (unbounded) probabilistic reachability of $\mathscr{F}_\perp$ from $s \in \mathscr{S}^+$:
\begin{align*}
	\Vp(s \sep \pi) := \mathbb{P} (
		s_{T} \in \mathscr{F}_\perp \, | \, s_0 = s, \, \pi )
\end{align*}
meaning the probability of reaching a failure state $\in \mathscr{F}_\perp$ at the terminal instant~$T$, given that an episode starts from the state $s_0 = s$ and follows the policy~$\pi$. $\Vp(s \sep \pi)$ thus quantifies how safe it is to follow the policy~$\pi$ from the initial state $s$. By definition, 
\begin{equation*}
    \Vp(s \sep \pi) = \indicator(s \in \mathscr{F}_\perp) \qquad \forall s \in \mathscr{S}_\perp
    %\label{eq:probabilistic reachability at failure states}
\end{equation*}
where $\indicator(\cdot)$ is the indicator function.

Given a safety threshold $\theta \in [0, 1)$, we define the sets of all safe and unsafe states, $\safeset(\theta \sep \pi)$ and $\unsafeset(\theta \sep \pi)$, respectively, as 
\begin{align*}
	\mathcal{S}(\theta \sep \pi) :=& \big \{ s \in \mathscr{S^+} \, \vert \, P(s \sep \pi) \leq \theta \big \}
	\\
	\mathcal{F}(\theta \sep \pi) :=& \big \{ s \in \mathscr{S^+} \, \vert \, P(s \sep \pi) > \theta \big \}
\end{align*}
The entire state space~$\mathscr{S^+}$ is then partitioned by these disjoint safe and unsafe regions as $\mathscr{S^+} = \mathcal{S}(\theta \sep \pi) \cup \mathcal{F}(\theta \sep \pi)$.

\paragraph{Value Functions} Given policy~$\pi$, define its value function~$V$ as
\[
	\Vr(s \sep \pi) := \mathbb{E} ( R_T \, | \, s_0 = s, \, \pi ).
\]
where $R_T$ is the return~\eqref{eq:return}. The value $\Vr(s \sep \pi)$ is a performance metric to be optimized at each $s \in \mathscr{S}$. At each terminal state $s \in \mathscr{S}_\perp$, it is directly given by $\Vr(s \sep \pi) = \mathscr{R}(s, \pi(s), s)$. Similarly, we define the action-value functions for each $(s, a) \in \mathscr{S^{+\!}} \times \mathscr{A}(s)$ as 
\begin{align*}
	\Qr(s, a \sep \pi) &:= \mathbb{E} ( R_T \, | \, s_0 a_0 = s a, \, \pi )
	\\
	\Qp(s, a \sep \pi) &:= \mathbb{P} (
		s_{T} \in \mathscr{F}_\perp \, | \, s_0 a_0 = s a, \pi )
\end{align*}
which are the same as $Q(s \sep \pi)$ and $P(s \sep \pi)$, except that they represent the value and probabilistic reachability, respectively, when the action~$a$ is taken at the initial state $s \in \mathscr{S}$ and then $\pi$ is followed.
% and thus induces partial order between the policies as $\pi \leq \pi'$ $\Longleftrightarrow$ $V($

% optimized, under the constraint w.r.t. $\Vp(s \sep \pi)$..

\section{Constrained Optimality}
\label{sec:constrained-optimality}

We denote $\optpolicy$ the assumed existent optimal policy that holds the following properties, labelled $\mathbf{P1}$---$\mathbf{P4}$, associated with threshold $\theta \in [0, 1)$. For notational convenience, we denote 
\begin{align*}
	\optsafeset(\theta) := \safeset(\theta \sep \optpolicy)
	\;\text{ and }\;
	\optunsafeset(\theta) := \unsafeset(\theta \sep \optpolicy)
\end{align*}
% \begin{align*}
% 	&\optsafeset(\theta) := \mathscr{S}(\theta \sep \optpolicy)
% 	\qquad\;\,
% 	\optunsafeset(\theta) := \mathscr{F}(\theta \sep \optpolicy)
%     \\
%     &\optVr{\theta}(s) := \Vr(s \sep \optpolicy)
%     \qquad
%     \optVp{\theta}(s) := \Vp(s \sep \optpolicy)
% \end{align*}
\begin{description}
	\item [$\mathbf{P1}$] \it $\optpolicy$ is uniformly optimal in the sense that for any policy~$\pi$,
    \begin{align}
		%\safeset(\theta \sep \pi) \supseteq \optsafeset(\theta) \; 
		P(s \sep \pi) \leq P(s \sep \optpolicy) \;
		\Longrightarrow \; V(s \sep \pi) \leq V(s \sep \optpolicy) \;\; \forall s \in \optsafeset(\theta)
		\label{eq:P1:performance}
		\\
		%\unsafeset(\theta \sep \pi) \supseteq \optunsafeset(\theta) \;
		V(s \sep \optpolicy) \leq V(s \sep \pi) \;
		\Longrightarrow \; P(s \sep \optpolicy) \leq P(s \sep \pi) \;\; \forall s \in \optunsafeset(\theta)
		\label{eq:P1:safety}
	\end{align}
\end{description}
In other words, $\mathbf{P1}$ means that $\optpolicy$ is Pareto optimal w.r.t. performance and safety, uniformly in its safe and unsafe regions $\optsafeset(\theta )$ and $\optunsafeset(\theta)$, respectively. 

There may exist multiple optimal policies that all satisfy $\mathbf{P1}$ but achieve different Pareto efficiency. The next property limits such optimality to the case(s) where the safety is maximally improved over the unsafe region.
\begin{description}
	\item [$\mathbf{P2}$] \it $\optpolicy$ is uniformly least unsafe over $\optunsafeset(\theta)$ in the sense that for any policy~$\pi$ s.t. $\pi = \optpolicy$ over $\optsafeset(\theta)$, 
    \begin{align}
		P(s \sep \optpolicy) \leq P(s \sep \pi) \qquad \forall s \in \optunsafeset(\theta)
		\label{eq:P2}
	\end{align}    
\end{description}
$\mathbf{P2}$ makes sense also in practice since we trade-off safe and performance within the safe region but \emph{not in the unsafe region}, in which safety comes to be the first priority to be optimized.

Next, it is desirable to have the following monotonicity property among the optimal policies w.r.t. different thresholds.
\begin{description}
	\item [$\mathbf{P3}$] \it If $0 \leq \vartheta \leq \vartheta' \leq 1$, then
		\begin{align}
			 	&V(s \sep \optpolicyvar ) \leq V(s \sep \optpolicyvarp) \quad \forall s \in \optsafeset(\vartheta')
			 	\notag
				\\
			 	&P(s \sep \optpolicyvar ) \leq \,P(s \sep \optpolicyvarp) \quad \forall s \in \mathscr{S}^+
			 \label{eq:monotonicity property for P}
		\end{align}
\end{description}
The intuition behind $\mathbf{P3}$ is that the weaker the constraint is (i.e., the larger $\theta$), the less conservative the optimal policy is (i.e., the larger probabilistic reachability, thus potentially the better performance). \Cref{eq:monotonicity property for P} means $\optpolicyvarp$ is less conservative than $\optpolicyvar$.
%, such that
% \begin{align*}
% 	\safeset(\vartheta' \sep \optpolicyvar) \supseteq \optsafeset(\vartheta')
% 	\;\text{ and }\;
% 	\unsafeset(\vartheta \sep \optpolicyvarp) \supseteq \optunsafeset(\vartheta)
% \end{align*}
In addition, we can see that if $\mathbf{P3}$ is true, then so are \eqref{eq:P1:performance} for $\theta = \vartheta'$ and $\pi = \optpolicyvar$, and \eqref{eq:P1:safety} for $\theta = \vartheta$ and $\pi = \displaystyle \optpolicyvarp$, in $\mathbf{P1}$. That is, the properties $\mathbf{P1}$ and $\mathbf{P3}$ are coherent.

For the next property, let the Bellman operator $\mathcal{T}_\theta$, on the space of functions from each $(s, a) \in \mathscr{S^{+\!}} \times  \mathscr{A}(s)$ to $\mathbb{R}^2$, be defined as
\[
    \mathcal{T}_\theta(Q, \mathscr{P}) := (Q', \mathscr{P}') 
\]
where $Q'(s, a) = \mathscr{R}(s, a, s)$ and $\mathscr{P}'(s, a) = \indicator(s \in \mathscr{F}_\perp)$ for all $s \in \mathscr{S}_\perp$, and otherwise,
\begin{align*}
    Q'(s, a) &= \mathbb{E} \big [ r_0 + \gamma \cdot Q(s_1, \pi'(s_1)) \, \big \vert \, s_0a_0 = sa \big ]
    \\
    \mathscr{P}'(s, a) &= \mathbb{E} \big [ \mathscr{P}(s_1, \pi'(s_1)) \, \big \vert \, s_0a_0 = sa \big ]
\end{align*}
for a policy $\pi'$ given by 
\begin{align}
    &\pi'(s) \in 
    \begin{cases}
        \displaystyle 
        \Argmax_{a \in \mathscr{A}_\theta(s \sep \mathscr{P})} Q(s, a) \;\, \text{ if } \mathscr{A}_\theta(s \sep \mathscr{P}) \neq \varnothing
        %\exists a \in \mathscr{A}(s) \text{ s.t. } \mathscr{P}(s, a) \leq \alpha
        \\
        \displaystyle
        \;\;\Argmin_{a \in \mathscr{A}(s)}\; \mathscr{P}(s, a) \;\, \text{ otherwise}
    \end{cases}
	\label{eq:pi prime}
	\\
	&\mathscr{A}_\theta(s \sep \mathscr{P}) := \{ a \in \mathscr{A}(s) \, \vert \, \mathscr{P}(s, a) \leq \theta \}
	\label{eq:constrained action set}
\end{align}
where 
$
\mathscr{A}_\theta(s \sep \mathscr{P})
$ is the constrained action set constructed from $\mathscr{P}$, and the dependency of $Q'$, $\mathscr{P}'$, and $\pi'$ on $\theta$ is implicit. 

In what follows, we state properties regarding the Bellman operator $\mathcal{T}_\theta$ and the optimal action-value functions denoted by
\[
    \optQr{\theta}(s, a) := \Qr(s, a \sep \optpolicy) \text{ and } \optQp{\theta}(s, a) := \Qp(s, a \sep \optpolicy)
\]
$\,$\\[-20pt]
\begin{description}
    \item [$\mathbf{P4}$] \it $(\optQr{\theta}, \optQp{\theta})$ is a fixed point of $\mathcal{T}_\theta$, i.e., 
    $
        (\optQr{\theta}, \optQp{\theta}) = \mathcal{T}_\theta (\optQr{\theta}, \optQp{\theta})
    $.\\[-5pt]
\end{description}
Note that $\mathbf{P4}$ is a reasonable property of optimality---if it is true, then the action $\hat a = \optpolicy(s)$ at state $s \in \optsafeset(\theta)$ yields the best value among those actions $a$ satisfying the constraint $\optQp{\theta}(s, a) \leq  \theta$ and at the unsafe state~$s \in \optunsafeset(\theta)$, $\hat a$ is safer than any other actions ${a \in \mathscr{A}(s)}$. The latter is also consistent with $\mathbf{P2}$, and $\mathbf{P4}$ is necessary for convergence of dynamic programming and reinforcement learning methods.

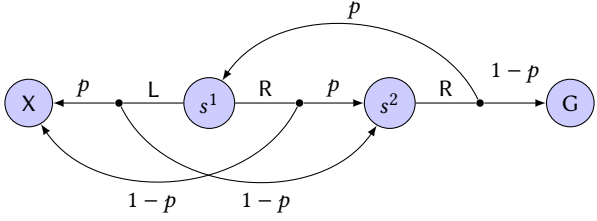
\begin{figure}
    \centering
    \begin{tikzpicture}
    [every node/.style={circle, minimum size = 2, inner sep = 1}]
    \node[fill](L1)at(1.2,0){};
    \node[fill](R1)at(3.6,0){};
    \node[fill](R2)at(6.0,0){};
    \draw[every node/.style={draw, circle, minimum size = 2em, fill=blue!20}]
    (0,0)node(X){\textsf{X}}
    (2.4,0)node(1){$s^1$}
    (4.8,0)node(2){$s^2$}
    (7.2,0)node(G){\textsf{G}};
    \draw[every edge/.style={->, >=latex, draw}]
    (1) edge[-]node[above]{\textsf{L}}(L1)
    (L1)edge node[above]{$p$}(X)
    (L1)edge[bend right=60]node[below right]{$1-p$}(2)
    (1) edge[-]node[above]{\textsf{R}}(R1)
    (R1)edge node[above]{$p$}(2)
    (R1)edge[bend left=60]node[below left]{$1-p$}(X)
    (2) edge[-]node[above]{\textsf{R}}(R2)
    (R2)edge node[above]{$1-p$}(G)
    (R2)edge[bend right=60]node[above]{$p$}(1);
    \end{tikzpicture}
    \caption{A simple counter-MDP for $\mathbf{P4}$ when $0.5 < p < 1$}
    \label{fig:counter-example}
\end{figure}

\section{Learning Instability}
\label{sec:pathological-behaviours}

In this section, we present a simple example---the ``counter-MDP'' shown in~\Cref{fig:counter-example}---which demonstrates that $\mathbf{P4}$ (the fixed point property) does not hold for certain values of threshold $\theta$. Dynamic programming is therefore unstable.

The state space of this example is $\mathscr{S^+} = \{ s^1, s^2, \mathsf{X}, \mathsf{G} \}$, with
$\mathscr{S} = \{ s^1, s^2 \}$,
$\mathscr{S}_\perp = \{ \mathsf{X}, \mathsf{G} \}$
and $\mathscr{F}_\perp = \{ \mathsf{X} \}$.
The action space $\mathscr{A^+} = \mathscr{A}(s^1) = \{\mathtt{L}, \mathtt{R}\}$, with $\mathscr{A}(s^2) = \{\mathtt{R}\}$.
For each transition during an execution there is a reward of $-1$.
Intuitively, the learning objective is to minimize the expected discounted path length, while keeping the probability of reaching $\mathsf{X}$ less than or equal to $\theta$. 
In general, the smaller~$\theta$, the higher the probability of the agent reaching $\mathsf{G}$.
    
We assume the initial state is $s^1$, where the agent chooses to move either to the left~(action $\Left$) or to the right~(action $\Right$). With probability $p > 0$ the agent moves in its chosen direction and with probability $1 - p$ it moves in the opposite direction. For simplicity, the agent has no choice in state $s^2$ and must choose action $\Right$. From $s^2$, the agent will reach the goal~$\mathsf{G}$ with probability $1-p$, or return to state $s_1$ with probability $p$. An episode terminates when the agent reaches a terminal state in $\mathscr{S}_\perp$.

Only two policies exist---choose $\Left$ in $s^1$ or choose $\Right$ in $s^1$---which we denote by $\pi_\Left$ and $\pi_\Right$, respectively. For notational simplicity, we denote the corresponding action-value functions in state $s^1$ by 
\begin{equation*}
	\begin{aligned}
		&\Qr_{a\Left} := \Qr(s^1, a \sep \pi_\Left)
		\\
		&\Qp_{a\Left} := \Qp(s^1, a \sep \pi_\Left)	
	\end{aligned}	
	\qquad
	\begin{aligned}
		&\Qr_{a\Right} := \Qr(s^1, a \sep \pi_\Right)
		\\
		&\Qp_{a\Right} := \Qp(s^1, a \sep \pi_\Right)	
	\end{aligned}
\end{equation*}
where the dependency on $s^1$ is implicit. Hence, for example, $\Qr_{a\Left}$ is the Q-function when the agent initially takes $a \in \{ \Left, \Right\}$ at $s^1$ and then follows $\pi_\Left$. 
Given the simplicity of the MDP, $\Qr_{a\Left}$ and $\Qp_{a\Left}$ can be given as explicit functions of probability $p$: 
\begin{align*}
	&\begin{aligned}
		&\Qp_\LL = \frac{p}{1 - pq}
		\\
		&\Qr_\LL = -\frac{1 + \gamma q}{1 - \gamma^2 pq}	
	\end{aligned}
	\qquad \qquad
	\begin{aligned}
		&\Qp_\RL = 1 - \frac{pq}{1 - pq}
		\\
		&\Qr_\RL = - \frac{1 + \gamma p + \gamma^2 p(p - q)}{1 - \gamma^2 pq}
	\end{aligned}	
\intertext{where $q := 1 - p$. Similarly,}
	&\begin{aligned}
		&\Qp_\LR = 2\cdot\frac{p}{p+1}
		\\
		&\Qr_\LR = -\frac{1+\gamma (1 - 2 p)}{1 - \gamma p}
	\end{aligned}
	\qquad\,
	\begin{aligned}
		&\Qp_\RR = \frac{1}{p+1}
		\\
		&\Qr_\RR = -\frac{1}{1 - \gamma p}
	\end{aligned}
\end{align*}
In~\Cref{fig:Q,fig:P} we plot these equations against $p$, for $\gamma=0.95$.
In the following analysis, to demonstrate the counter-example, we consider only $0.5 < p < 1$.
\begin{figure} [h]
    %   \begin{subfigure}{0.47\columnwidth}
    %      \adjincludegraphics[clip, width=\textwidth]{Figures/QL}   
    %      \caption{$\Qr_{a\Left}$ vs $p$}
    %       \label{fig:QL}
    %   \end{subfigure}
    %   \begin{subfigure}{0.51\columnwidth}
    %      \adjincludegraphics[clip, width=\textwidth]{Figures/QR}          
    %      \caption{$\Qr_{a\Right}$ vs $p$}
    %       \label{fig:QR}
    %   \end{subfigure}
    \begin{subfigure}{0.49\columnwidth}
    \begin{tikzpicture}
    \pgfmathsetmacro{\gamma}{0.95}  % discount factor
    \begin{axis}[
        width = 1.2\textwidth,
        axis lines = left,
        xlabel = $p$,
        xlabel style = {font=\small, yshift=3pt},
        % ylabel = {$\Qr$-value},
        % ylabel style = {font=\small},
        % ylabel shift = -10pt,
        style = {font=\scriptsize},
        legend pos = {south east},
        legend style={draw=none, fill=none, at={(1,0.3)}},
        legend image post style={xscale=0.5},
        domain=0.5:1,
        ymin =-3,
        xtick = {0, 0.5, 1},
        ytick = {-3, -1}
    ]
    \addplot[color=orange, line width=0.6pt]
        {(1+\gamma*(1-x))/(\gamma*\gamma*x*(1-x)-1)};
    \addlegendentry{$\Qr_\LL$}
    \addplot[color=blue, line width=0.6pt]
        {(1+\gamma*x+\gamma*\gamma*x*(2*x-1))/(\gamma*\gamma*x*(1-x)-1)};
    \addlegendentry{$\Qr_\RL$}
    \draw[densely dotted](0.5,-3)--(0.5,-1.904);
    \end{axis}
    \end{tikzpicture}
    \caption{$\Qr_{a\Left}$ vs $p$}
    \label{fig:QL}
    \end{subfigure}
    \begin{subfigure}{0.49\columnwidth}
    \begin{tikzpicture}
    \pgfmathsetmacro{\gamma}{0.95}  % discount factor
    \begin{axis}[
        width = 1.2\textwidth,
        axis lines = left,
        xlabel = $p$,
        xlabel style = {font=\small, yshift=3pt},
        % ylabel = {$\Qr$-value},
        % ylabel style = {font=\small},
        % ylabel shift = -10pt,
        style = {font=\scriptsize},
        legend style={draw=none, fill=none, at={(1.0,0.8)}, anchor=north east},
        legend image post style={xscale=0.5},
        domain=0.5:1,
        xtick = {0, 0.5, 1},
        ytick = {-log10(20), -1, -log10(2), 0},
        yticklabels = {$-20$, $-10$, $-2$, $-1$}
    ]
    \addplot[smooth, color=orange, line width=0.6pt]{-log10((1+\gamma*(1-2*x))/(1-\gamma*x))};
    \addlegendentry{$\Qr_\LR$}
    \addplot[color=blue, line width=0.6pt]{-log10(1/(1-\gamma*x))};
    \addlegendentry{$\Qr_\RR$}
    % \draw[densely dotted](0.5,-1.3)--(0.5,-0.28);
    \end{axis}
    \end{tikzpicture}
    \caption{$\Qr_{a\Right}$ vs $p$}
    \label{fig:QR}
    \end{subfigure}
    \caption{Q-functions for $s^1$ w.r.t. (a) $\pi_\Left$ and (b) $\pi_\Right$, for $\gamma = 0.95$}
    \label{fig:Q}
\end{figure}
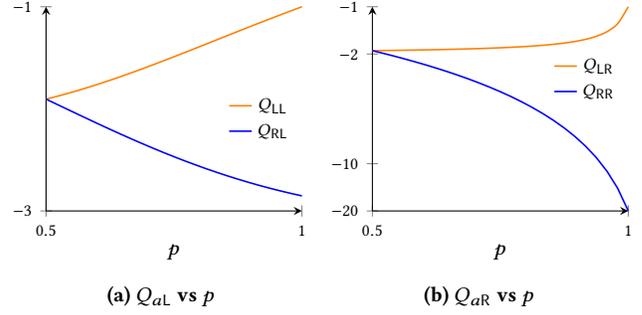
\begin{figure} [h]
    %   \begin{subfigure}{0.49\columnwidth}
    %      \adjincludegraphics[clip, width=\textwidth]{Figures/PL}   
    %      \caption{$\Qp_{a\Left}$ vs $p$}
    %       \label{fig:PL}
    %   \end{subfigure}
    %   \begin{subfigure}{0.49\columnwidth}
    %      \adjincludegraphics[clip, width=\textwidth]{Figures/PR}          
    %      \caption{$\Qp_{a\Right}$ vs $p$}
    %       \label{fig:PR}
    %   \end{subfigure}
    \begin{subfigure}{0.49\columnwidth}
    \begin{tikzpicture}
    \begin{axis}[
        width = 1.2\textwidth,
        axis lines = left,
        xlabel = $p$,
        xlabel style = {font=\small, yshift=3pt},
        % ylabel = {$\mathscr{P}$-value},
        % ylabel style = {font=\small, xshift=-5pt},
        % ylabel shift = -10pt,
        style = {font=\scriptsize},
        legend pos = {south east},
        legend style={draw=none, fill=none, at={(1.04,0)}},
        legend image post style={xscale=0.5},
        domain=0.5:1,
        ymin=0.5,
        xtick = {0, 0.5, 0.7, 1},
        ytick = {0.5, 0.85, 1}
    ]
    \addplot[color=orange, line width=0.6pt]{x/(1-x*(1-x))};
    \addlegendentry{$\mathscr{P}_\LL$}
    \addplot[color=blue, line width=0.6pt]{1-x*(1-x)/(1-(x*(1-x))};
    \addlegendentry{$\mathscr{P}_\RL$}
    \addplot[color=orange, densely dashed, line width=0.6pt]{2*x/(x+1)};
    \addlegendentry{$\mathscr{P}_\LR$}
    \addplot[mark=*, mark size=0.8pt] coordinates {(0.7, 0.85)};
    \draw[dotted, line width=0.6pt](0.7,0)--(0.7,0.85)--(0,0.85);
    % \draw[densely dotted](0.5,0)--(0.5,2/3);
    \end{axis}
    \end{tikzpicture}
    \caption{$\mathscr{P}_{a\Left}$ vs $p$}
    \label{fig:PL}
    \end{subfigure}
    \begin{subfigure}{0.49\columnwidth}
    \begin{tikzpicture}
    \begin{axis}[
        width = 1.2\textwidth,
        axis lines = left,
        xlabel = \(p\),
        xlabel style = {font=\small, yshift=3pt},
        % ylabel = {$\mathscr{P}$-value},
        % ylabel shift = -5pt,
        % ylabel style = {font=\small},
        style = {font=\scriptsize},
        legend style={draw=none, fill=none, at={(1.0,0.6)}, anchor=north east},
        legend image post style={xscale=0.5},
        domain=0.5:1,
        ymin=0.5,
        xtick = {0, 0.5, 1},
        ytick = {0.5, 1}
    ]
    \addplot[color=orange, line width=0.6pt]{2*x/(x+1)};
    \addlegendentry{$\Qp_\LR$}
    \addplot[color=blue, line width=0.6pt]{1/(x+1)};
    \addlegendentry{$\Qp_\RR$}
    \draw[densely dotted](0.5,0)--(0.5,2/3);
    \end{axis}
    \end{tikzpicture}
    \caption{$\Qp_{a\Right}$ vs $p$}
    \label{fig:PR}
    \end{subfigure}
      \caption{P-functions for $s^1$ w.r.t. (a) $\pi_\Left$ and (b) $\pi_\Right$}
      \label{fig:P}
\end{figure}
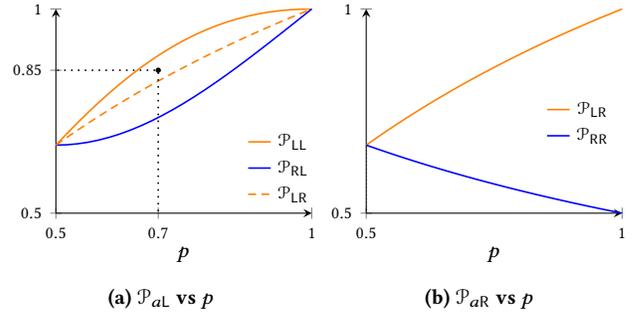

Figures~\ref{fig:QL} and \ref{fig:QR} plot the Q-values of each action in state $s^1$, given the policy is $\pi_\Left$ and $\pi_\Right$, respectively.
We see that regardless of the policy and the value of $p > 0.5$, action~$\Left$ always has the greater Q-value.
Hence, while the actual Q-value depends on the policy, there is never any ambiguity about which action to choose to maximize the reward:
if the constraint is satisfied, the agent should always choose action~$\Left$.

Figures~\ref{fig:PL} and \ref{fig:PR} plot the $\mathscr{P}$-values of each action in state $s^1$, given the policy is $\pi_\Left$ and $\pi_\Right$, respectively.
As in the case of the Q-values, we see that action~$\Left$ for $p > 0.5$ unambiguously has greater $\mathscr{P}$-value, with the actual $\mathscr{P}$-value depending on the policy.
In contrast to the Q-value case, however, the probability that the $\mathscr{P}$-value represents must also satisfy the constraint, i.e., be less than or equal to threshold $\theta$.
The relative performance of the two actions $\Left$ and $\Right$ is therefore not sufficient to decide which one is optimal.

To make this decision, we first note that the $\mathscr{P}$-value for action~$\Left$ under $\pi_\Left$ ($\Qp_\LL$) is the \emph{true} probability for taking action~$\Left$, and the $\mathscr{P}$-value for action~$\Right$ under $\pi_\Right$ ($\Qp_\RR$) is the true probability for taking action~$\Right$.
We then call the $\mathscr{P}$-value for action~$\Left$ under $\pi_\Right$ ($\Qp_\LR$) the \emph{estimated} probability for taking action~$\Left$ (under $\pi_\Right$), and call the $\mathscr{P}$-value for action~$\Right$ under $\pi_\Left$ ($\Qp_\RL$) the estimated probability of action~$\Right$ (under $\pi_\Left$).

In~\Cref{fig:PL} we see that the true probability of action~$\Left$ is always greater than the estimated probability of action~$\Left$.
If we choose a threshold $\theta$ between the true and estimated values, such that $\Qp_\LL > \theta > \Qp_\LR$, we find that a simple learning agent will not be able to decide which action is optimal using only the $\mathscr{P}$-values.
Suppose, during the learning process, the current policy is $\pi_\Left$, which we know maximizes the Q-value, the agent will see that the $\mathscr{P}$-value ($\Qp_\LL$) is greater than $\theta$, which does not satisfy the constraint.
\Cref{fig:PL} shows that the estimated value of action~$\Right$ under $\pi_\Left$ ($\Qp_\RL$) will satisfy the constraint, so the agent chooses $\pi_\Right$.
Under $\pi_\Right$, however, the estimated value of action~$\Left$ ($\Qp_\LR$) appears now to satisfy the constraint, so the (memoryless) agent chooses $\pi_\Left$ once again.

To see this phenomenon in a concrete learning example, consider the policy iteration process described in \Cref{table:unstable example}, with $(p, \theta, \gamma) = (0.7, 0.85, 0.95)$ and $\pi_\Right$ as its initial policy. At the first iteration~$i = 1$, $\mathscr{P}_\LR \leq \theta$ indicates that action~$\Left$ seems to be safe, and thus we choose $\pi_\mathsf{L}$. However, at the next iteration ($i = 2$), $\mathscr{P}_\LL \nleq \theta$ shows that $\pi_\Left$ is \emph{not} safe, which forces us to choose $\pi_\Right$ again. Unfortunately, $\mathscr{P}_\LR \leq \theta$ at iteration $i = 2$ falsely indicates that action~$\Left$ is safe, again! This oscillation continues ad infinitum.

In summary, we see that a naive learning approach with the example shown in~\Cref{fig:counter-example} will not reach a fixed point, so $\mathbf{P4}$ does not hold in general.

\iffalse
\todo{(for details on the their calculations and more, see Appendix~?).}
\todo{
\begin{enumerate}
	\item Say about details on why $\mathbf{P4}$ is violated and for which (range of) $\theta$... In the simplest case, we can just [and must] mention the case of $(p, \theta, \gamma) = (0.7, 0.85, 0.95)$. Also note that Figures 2 and 3 uses $\gamma = 0.95$ and I will use the same $\gamma$ (and same $p = 0.7$ and $\theta = 0.85$ in the next section).
	\item Write ``Algorithm~1. Naive Value Iteration'' (and Naive Policy Iteration, too?). Although we gave a policy iteration example, we need value iteration experiments instead, since VI exactly implements fixed point iteration of $\mathcal{T}_\theta$ whereas PI does not.
	\item Implement naive value iteration for the MDP, simulate for all $\theta$, and show that it does not converge... (e.g., at $\theta = 0.85$).
	\item Write down Appendix~?? (if we have time).
\end{enumerate}
}

\todo{Pathological behaviour depends on the threshold, (degree of) randomizations of the environment, and what?}

\fi

\newcommand{\violation}{\red{\Qp_\LL \nleq \theta\xspace}}
\newcommand{\satisfaction}{\Qp_\LR \leq \theta\xspace}
\newcommand{\satisfactionRL}{\Qp_\RL \leq \theta\xspace}
\newcommand{\satisfactionRR}{\Qp_\RR \leq \theta\xspace}

\begin{table*} [t!]
	\centering
	\caption{Policy iteration on the MDP in \Cref{fig:counter-example} for $(p, \theta, \gamma) = (0.7, 0.85, 0.95)$}
	\label{table:unstable example}
	\begin{tabular}	{c|c|c|c|c|c|c|c}
			\multicolumn{2}{c|}{Iteration~$i$} & $1$ & $2$ & $3$ &	$4$ & $5$ & $\qquad\cdots\qquad$ \\
			\multicolumn{2}{c|}{Given policy} & $\pi_\Right$ & $\pi_\Left$ & $\pi_\Right$ & $\pi_\Left$ & $\pi_\Right$ & $\qquad\cdots\qquad$ \\
			\hline
			\multirow{2}{*}{\makecell{\vspace{-6pt}\\Constraints}} \bigstrut & $\Left$ & $\Qp_\LR\approx0.82 \leq \theta=0.85$ & $\red{\Qp_\LL\approx0.89 \nleq \theta}$ & $\satisfaction$ & $\violation$ & $\satisfaction$ & $\qquad\cdots\qquad$ 
			\\
			   \bigstrut & $\Right$ & $\Qp_\RR\approx0.59 \leq \theta=0.85$ & $\Qp_\RL\approx0.73 \leq \theta$ & $\satisfactionRR$ & $\satisfactionRL$ & $\satisfactionRR$ & $\qquad\cdots\qquad$
	\end{tabular}
\end{table*}

\newcommand{\Constr}[1]{\mathrm{C}_{#1}}

\begin{table*} [ht]
	\centering
    \caption{Policy iteration  with recursive constraints on the MDP in \Cref{fig:counter-example} for $(p, \theta, \gamma) = (0.7, 0.85, 0.95)$}
	\label{table:stable example}
	\begin{tabular}	{c|c|c|c|c|c|c}
			\multicolumn{2}{c|}{Iteration~$i$} & $1$ & $2$ & $3$ &	$4$ & $\qquad\cdots\qquad$ \\
			\multicolumn{2}{c|}{Given policy} & $\pi_\Right$ & $\pi_\Left$ & $\pi_\Right$ & $\pi_\Right$ & $\qquad\cdots\qquad$ \\
			\hline
			\multirow{2}{*}{\makecell{\vspace{-6pt}\\Constraints}} \bigstrut & $\Left$ & $\Constr{\Left} \gets (\satisfaction)$ & $\red{\Constr{\Left}} \gets (\violation) \wedge \Constr{\Left}$ & $\red{\Constr{\Left}} \gets (\satisfaction) \wedge \red{\Constr{\Left}}$ & $\red{\Constr{\Left}} \gets (\satisfaction) \wedge \red{\Constr{\Left}}$ & $\qquad\cdots\qquad$ 
			\\
			\bigstrut  & $\Right$ & $\Constr{\Right} \gets (\satisfactionRR)$ & $\Constr{\Right} \gets (\satisfactionRL) \wedge \Constr{\Right}$ & $\Constr{\Right} \gets (\satisfactionRR) \wedge \Constr{\Right}$ & $\Constr{\Right} \gets (\satisfactionRR) \wedge \Constr{\Right}$ & $\qquad\cdots\qquad$
	\end{tabular}	
\end{table*}

\section{Recursive Constraints}
\label{sec:recursive-constraints}

In this section, we answer to the following questions---(i) ``what is wrong with fixed point property~$\mathbf{P4}$?''; (ii) ``how can we solve it?'' 

To address the former, we point out the mismatch between $\mathbf{P1}$ and $\mathbf{P4}$. If $\optpolicy$ holds \eqref{eq:P1:performance} in $\mathbf{P1}$, then it satisfies
\begin{align}
    &\optpolicy(s) \in \Argmax_{a \in \optA{\theta}(s)} \optQr{\theta}(s, a)
    \qquad \forall s \in \optsafeset(\theta)
    \label{eq:argmax from P1}
    \\
    &\mathscr{\hat A}_{\theta}(s) := \{ a \in \mathscr{A}(s) \, \vert \, \optQp{\theta}(s, a) \leq \Vp(s \sep \optpolicy) \}
    \notag 
\end{align}
Intuitively, \eqref{eq:argmax from P1} means that $\optpolicy$ must yield a higher value over the safe region, $\optsafeset(\theta)$, than any of its conservative one-point modifications in $\optsafeset(\theta)$. However, we can easily notice the difference between the two constrained action sets---$\mathscr{\hat A}_{\theta}(s)$ in \eqref{eq:argmax from P1} and $\mathscr{A}_{\theta}(s \sep \optQp{\theta})$, where $\mathscr{A}_{\theta}(s \sep \cdot )$ is defined by \eqref{eq:constrained action set} and used to construct the policy~\eqref{eq:constrained action set} and thus the Bellman operator~$\mathcal{T}_\theta$ in $\mathbf{P4}$. Here, $\optQp{\theta}(s, a)$ is constrained by the threshold $\theta$ in the latter but by $P(s \sep \optpolicy)$ in the former. For each $s \in \optsafeset(\theta)$, since $P(s \sep \optpolicy) \leq \theta$ holds by the definition of $\optsafeset(\theta)$, the former $\mathscr{\hat A}_{\theta}(s)$ is more conservative than the latter $\mathscr{A}_{\theta}( s \sep \optQp{\theta})$, that is, $\mathscr{\hat A}_{\theta}(s) \subseteq \mathscr{A}_{\theta}(s \sep \optQp{\theta})$. Therefore, $\mathbf{P1} \centernot \Longrightarrow \mathbf{P4}$, in general.

In a similar manner to~\eqref{eq:argmax from P1}, $\optpolicy$ is safer over $\optunsafeset(\theta)$ than any of its better-performing one-point modifications in $\optunsafeset(\theta)$ if it satisfies \eqref{eq:P1:safety} in $\mathbf{P1}$. In this case, we have
\begin{align}
    &\optpolicy(s) \in \Argmin_{a \in \optAcinarg{\theta}(s)} \optQp{\theta}(s, a)
    \qquad \forall s \in \optunsafeset(\theta)
    \label{eq:argmin from P1}
    \\
    &\optAc{\theta}(s) := \{ a \in \mathscr{A}(s) \, \vert \, \optQr{\theta}(s, a) \geq \Vr(s \sep \optpolicy) \}
    \notag
\end{align}
On the other hand, if $\optpolicy$ holds $\mathbf{P2}$, then it satisfies \eqref{eq:argmin from P1} with $\optAc{\theta}(s)$ replaced by $\mathscr{A}(s)$ shown in the policy~\eqref{eq:pi prime} on the unsafe $\mathrm{argmin}$-part. Therefore, from the standard dynamic programming theory, we can conclude that $\mathbf{P1}$ and $\mathbf{P2}$ imply $\mathbf{P4}$ if the constrained action set $\mathscr{A}_{\theta}( s \sep \optQp{\theta})$ is equal to or replaced by $\mathscr{\hat A}_\theta(s)$. 

From the discussions above on the conservatism of $\mathbf{P1}$ w.r.t. $\mathbf{P4}$, we hypothesize that 
\begin{description}
	\item \emph{$\mathscr{A}_{\theta}( \cdot \sep \optQp{\theta})$ has to be replaced with a constrained action set, e.g., $\optA{\theta}(\cdot)$, that is more conservative, for $\mathbf{P4}$ to be true.}
\end{description}
In fact, the hypothesis is true for the counter-MDP in \Cref{fig:counter-example}. To see this, we revisit the policy iteration example with $(p, \theta, \gamma) = (0.7, 0.85, 0.95)$ and $\pi_\Right$ as its initial policy, but also with recursive constraints illustrated in Table~\ref{table:stable example}. Here, the meaning of the recursive constraints is clear from Table~\ref{table:stable example}---for each action~$a \in \{ \Left, \Right\}$, we superimpose all the constraints on $a$ by recursion, up to the current iteration, and use it to judge whether $a$ is a safe action or not. Denoting $\Constr{a}(i)$ such a recursive constraint for action~$a$, made at iteration $i$, then at each iteration in \Cref{table:stable example}, it satisfies by its construction
\begin{align*}
	&\Constr{\Left}(1) = (\satisfaction)
	\\
	&\red{\Constr{\Left}(2)} = (\violation) \wedge \Constr{\Left}(1) = (\violation) \wedge (\satisfaction)
	\\	
	&\red{\Constr{\Left}(3)} = (\satisfaction) \wedge \red{\Constr{\Left}(2)} = (\satisfaction) \wedge (\violation)
	\\
	&\red{\Constr{\Left}(4)} = (\satisfaction) \wedge \red{\Constr{\Left}(3)} = (\satisfaction) \wedge (\violation)
	\\
	&\quad \red{\vdots}  \qquad\;\,\qquad \vdots
	\qquad\qquad \red{\vdots}
	\qquad\;\,\qquad \vdots \qquad\qquad\quad \red{\vdots}
\end{align*}
From this, we observe that the constraint $\Constr{\Left}$ on the action~$\Left$ at $s^1$ is now stabilized and thereby, yields the same safe policy~$\pi_\Right$ from iteration~$i = 3$, as shown in \Cref{table:stable example}, whereas it was not without such recursive constraints, as illustrated in \Cref{table:unstable example}. 

On the other hand, the idea of recursive constraints, demonstrated in Table~\ref{table:stable example}, still has the issue that except for policy iteration, a dynamic programming or reinforcement learning method typically does not wait until its value function (e.g., $\mathscr{P}(s, a \sep \pi)$) is accurately estimated. In Table~\ref{table:stable example} for example, if $\Qp_\LL$ and/or $\Qp_\LR$ is not correctly estimated at the previous iterations, then the recursive constraint $\Constr{\Left}$ at the current iteration can be so deteriorated and messy as it is constructed from all the previous constraints, including those based on inaccurate predictions at the early stages. In particular, the initial values of $\Qp_\LL$ and $\Qp_\LR$ are typically random and has no information, which introduces and transfers absurd random constraints, to all iterations. Hence, the recursive constraints at and around the initial stages must be also stabilized, in order to generalize themselves to a broad class of reinforcement learning methods.

In order to solve such a remaining issue on stability, we (i) replace the iteration axis in \Cref{table:stable example} with the axis of horizon window ${n = 1,2,\dots, N}$ %, as shown in \Cref{table:stable example with horizon}, 
 and (ii) replace the constraint at stage~$n$ 
%at stage~$n$ (which is now at horizon $n$ in the new axis) 
with 
\[
    \textstyle \max_{m \in \nrange{1}{n}} \optQp{}^m(s, a) \leq \theta
\] 
where $\optQp{}^n(s, a)$ is for an (over-)approximation of the $n$-bounded probabilistic reachability 
\begin{align*}
	\Qp^n(s, a \sep \pi) &:= \mathbb{P} (
		s_{\min(T, n)} \in \mathscr{F}_\perp \, | \, s_0 a_0 = s a, \pi )	
\end{align*}
w.r.t. the policy~$\pi = {\hat \pi}^{n-1}$ obtained at the previous stage $n - 1$. Here, an over-approximation means $0 \leq \Qp^n(s, a \sep {\hat \pi}^{n-1}) \leq \optQp{}^n(s, a) \leq 1$. 
%rather than the unbounded one $\mathscr{P}(s, a \sep \pi) \leq \theta$. 
To describe our proposal, we also denote 
\begin{align*}
	\Vp^n(s \sep \pi) &:= \mathbb{P} (
		s_{\min(T, n)} \in \mathscr{F}_\perp \, | \, s_0 = s, \pi )	
\end{align*}
which satisfies $\Vp^n(s \sep \pi) = \Qp^n(s, \pi(s) \sep \pi)$ for all $s \in \mathscr{S}^+$.

Note that $\Qp^n(\cdot \sep \pi)$ at horizon $n = 1$ is now stable since it does not depend on the policy anymore, as shown below:
\begin{equation*}
	\optQp{}^1(s, a) := \Qp^1(s, a \sep \pi ) = \mathbb{P} (s_1 \in \mathscr{F}_\perp | \, s_0 a_0 = s a )	
	% \quad \forall s \in \mathscr{S}
\end{equation*}
From $\optQp{}^1(\cdot)$, we construct the first policy ${\hat \pi}^1$ on the horizon axis as (in this case, substitute $n = 1$)
\begin{align}
	\begin{aligned}
    &{\hat \pi}^n(s) \in 
    \begin{cases}
        \displaystyle 
        \Argmax_{a \in \optA{}^n(s)} \optQr{}^n(s, a) \;\, \text{ if } \optA{}^n (s) \neq \varnothing
        %\exists a \in \mathscr{A}(s) \text{ s.t. } \mathscr{P}(s, a) \leq \alpha
        \\
        \displaystyle
        \Argmin_{a \in \mathscr{A}(s)}\; \optQp{}^n(s, a) \;\, \text{ otherwise}
    \end{cases}
    \\
    &\optA{}^n(s) := \{ a \in \mathscr{A}(s) \, \vert \, \Constr{a}(n \sep s) \}
    \qquad \optQr{}^n(s, a) := \Qr{}(s, a \sep {\hat \pi}^n )
    \end{aligned}
    \label{eq:construction of pi^n}
\end{align}
where the constraint $\Constr{a}(1 \sep s) := (\optQp{}^1(s, a) \leq \theta)$, and the dependency on the threshold $\theta$ is all implicit. Also note that the horizons of $\optQr{}^1$ and $\optQp{}^1$ are $\infty$ and $1$, respectively.

At the next stage $n = 2$, note that the $n$-bounded probabilistic reachability w.r.t. ${\hat \pi}^1$ satisfies the Bellman equation of the form
\begin{align}
	\Qp{}^2(s, a \sep {\hat \pi}^1) 
	= \mathbb{E} \big [ \optQp{}^1(s_1, {\hat \pi}^1(s_1) ) \, \big \vert \, s_0a_0 = sa \big ]
	% \quad\quad \forall s \in \mathscr{S}	
	\label{eq:BE for P^2}
\end{align}
Therefore, letting $\optQp{}^2(s, a) := \Qp{}^2(s, a \sep {\hat \pi}^1)$, the second policy~$\hat \pi^2$ can be easily constructed via \eqref{eq:construction of pi^n}, whose constraint is recursively defined as $\Constr{a}(2 \sep s) := (\optQp{}^2(s, a) \leq \theta) \wedge \Constr{a}(1 \sep s)$. 
\iffalse
Here, also note that $\optQp{}^2 \geq \optQp{}^1$ since $\optQp{}^1(s_1, a ) = 1$ $\forall a \in \mathscr{A}(s_1)$ if $s_1 \in \mathscr{F}_\perp$, hence the Bellman equation~\eqref{eq:BE for P^2} implies 
\begin{align*}
		\optQp{}^2(s, a) 
	&= \mathbb{P} \big [ (s_1 \in \mathscr{F}_\perp) \vee (s_2 \in \mathscr{F}_\perp) \, \big \vert \, s_0a_0 = sa, \, {\hat \pi}^1 \big ]
	\\
	&\geq \mathbb{P} \big [ s_1 \in \mathscr{F}_\perp \, \big \vert \, s_0a_0 = sa \big ]
\end{align*}
\fi

At horizon $n =  3, 4, 5, \dots, N$, the $n$-bounded probabilistic reachability $\Qp{}^n(\cdot \sep {\hat \pi}^{n-1})$ w.r.t. the policy ${\hat \pi}^{n-1}$ given at the previous step $n-1$ satisfies the Bellman equation: 
\begin{align*}
	\Qp{}^n(s, a \sep {\hat \pi}^{n-1}) 
	= \mathbb{E} \big [ \Vp^{n-1}(s_1 \sep {\hat \pi}^{n-1}) \, \big \vert \, s_0a_0 = sa \big ]
	%\qquad \forall s \in \mathscr{S}
\end{align*}
However, in order to obtain $\Vp^{n-1}(\cdot \sep {\hat \pi}^{n-1})$, we need to calculate $\Vp^{m}(\cdot \sep {\hat \pi}^{n-1})$ and use it in the backward induction for $\Vp^{m+1}(\cdot \sep {\hat \pi}^{n-1})$, all the way through $m = 1, 2, 3, \dots, n-1$. The longer the horizon $n$ is, the more complexity this procedure induces in space and time. Instead, our design choice is to use a substitute $\optQp{}^{n-1}(s, {\hat \pi}^{n-1}(s))$ obtained at the previous stage~$n-1$. Therefore, we define 
\begin{align*}
	\optQp{}^n(s, a) 
	:= \mathbb{E} \big [ \optQp{}^{n-1}(s_1 \sep {\hat \pi}^{n-1}(s_1)) \, \big \vert \, s_0a_0 = sa \big ]
	%\quad \forall s \in \mathscr{S}
\end{align*}
and construct the policy ${\hat \pi}^n$ at the current horizon~$n$ via \eqref{eq:construction of pi^n}, w.r.t. the recursive constraint 
\[
	\Constr{a}(n \sep s) := (\optQp{}^n(s, a) \leq \theta) \wedge \Constr{a}(n-1 \sep s)
\]
Here, $\optQp{}^{n-1}(s, {\hat \pi}^{n-1}(s))$ typically over-approximates $\Vp^{n-1}(s \sep {\hat \pi}^{n-1})$ since ${\hat \pi}^{n-1}$ is constructed with a fewer constraints than ${\hat \pi}^{n-2}$, and 
\[
	\begin{cases}
		\;\;\Vp^{n-1}(s \sep {\hat \pi}^{n-1}) \;\, = \Qp^{n-1}(s, {\hat \pi}^{n-1}(s) \sep \red{{\hat \pi}^{n-1}})
		\\
		\optQp{}^{n-1}(s, {\hat \pi}^{n-1}(s)) = \Qp^{n-1}(s, {\hat \pi}^{n-1}(s) \sep \red{{\hat \pi}^{n-2}})
	\end{cases}
\]

Finally, we provide the policy~$\pi^N$ at the last horizon $N$ as the receding-horizon solution that is potentially conservative but  able to uniformly and {(sub-)optimally} improve the performance subject to $N$-bounded probabilistic reachability constraint imposed on every state. To address the instability issue, the final policy~$\pi^N$ has the recursive constraints $\Constr{a}(N \sep s)$ that contain all constraints w.r.t. shorter horizons, i.e., 
\[
	\Constr{a}(N \sep s) = \bigwedge_{n \in \nrange{1}{N}} (\optQp{}^n(s, a) \leq \theta) \Longleftrightarrow \Big ( \max_{n\in \nrange{1}{N}} \optQp{}^n(s, a) \Big ) \leq \theta
\]
where each $\optQp{}^n(\cdot)$ is recursively defined from the initial one $\optQp{}^{1}(\cdot)$ that is independent of any policy and hence can be stably obtained. 

Since the recursive constraint $\Constr{a}(N \sep s)$ makes the constrained action set $\optA{}^N(s)$ monotonically decreasing as $N$ increases, and since $\optA{}^N(s)$ is finite, it is stabilized within a finite horizon, say $M$, after which the process becomes the same as that for unconstrained MDPs but w.r.t. the action space $\optA{}^M(\cdot)$. That is, the final policy~$\pi^N$ converges as $N \to \infty$, at the infinite-horizon.

\pgfplotstableset{col sep=comma}
\begin{figure*}[ht]
\centering
\begin{minipage}{0.49\textwidth}
\pgfplotstableread{Naive.csv}\naivedata
\begin{subfigure}{0.49\columnwidth}
\begin{tikzpicture}
%threshold,P-values-est,P-values-true,P-values-horizon-1,threshold,V-values-est,V-values-true
\begin{axis}[
    axis lines=left,
    width=1.2\textwidth,
    xlabel={$\theta$},
    ymin=0.0,
    ymax=1.0,
    xmin=0.0,
    xmax=1.0,
    tick pos=left,
    domain=0:1,
    legend pos={north west},
    legend style={draw=none, fill=none, font=\sffamily\small},
    legend cell align={left},
    legend image post style={xscale=0.5},
    style = {font=\scriptsize},
    xlabel style = {font=\small},
]
    \addplot[color=gray, densely dashed, line width=1.1pt]{x};
    \addlegendentry{threshold}
    \addplot[color=blue, line width=1.1pt] table [
    x = threshold,
    y = P-values-true,
    ]{\naivedata};
     \addlegendentry{true}
    \addplot[color=orange, line width=1.1pt] table [
    x = threshold,
    y = P-values-est,
    ]{\naivedata};
     \addlegendentry{estimate}
\end{axis}
\end{tikzpicture}
         \caption{$\Vp(s_0 \sep {\hat \pi})$ vs $\theta$}
\end{subfigure}
\begin{subfigure}{0.49\columnwidth}
\begin{tikzpicture}
%threshold,P-values-est,P-values-true,P-values-horizon-1,threshold,V-values-est,V-values-true
\begin{axis}[
    width=1.2\textwidth,
    xlabel={$\theta$},
    axis lines=left,
    xmin=0.0,
    xmax=1.0,
    ymin=-2.0,
    ymax=-1.4,
    tick pos=left,
    legend pos={north west},
    legend style={draw=none, fill=none, font=\sffamily\small},
    legend cell align={left},
    legend image post style={xscale=0.5},
    style = {font=\scriptsize},
    xlabel style = {font=\small}
]
    \addplot[color=blue, line width=1.1pt] table [
    x = threshold,
    y = V-values-true,
    ]{\naivedata};
     \addlegendentry{true}
    \addplot[color=orange, line width=1.1pt] table [
    x = threshold,
    y = V-values-est,
    ]{\naivedata};
     \addlegendentry{estimate}
\end{axis}
\end{tikzpicture}
         \caption{$\Vr(s_0 \sep {\hat \pi})$ vs $\theta$}
\end{subfigure}
%      \begin{subfigure}{0.49\columnwidth}
%         \adjincludegraphics[clip, width=\textwidth]{Figures/Naive_VI_P.png}
%         \caption{$\Vp(s_0 \sep {\hat \pi})$ vs $\theta$}
%      \end{subfigure}
%      \begin{subfigure}{0.49\columnwidth}
%         \adjincludegraphics[clip, width=\textwidth]{Figures/Naive_VI_Q.png}
%         \caption{$\Vr(s_0 \sep {\hat \pi})$ vs $\theta$}
%      \end{subfigure}
      \caption{Experimental results for naive value iteration}
      \label{fig:NaiveVI}
\end{minipage}
\hspace{0.5em}
\begin{minipage}{0.49\textwidth}
\pgfplotstableread{Recursive.csv}\recursivedata
\begin{subfigure}{0.49\columnwidth}
\begin{tikzpicture}
%threshold,P-values-est,P-values-true,P-values-horizon-1,threshold,V-values-est,V-values-true
\begin{axis}[
    axis lines=left,
    width=1.2\textwidth,
    xlabel={$\theta$},
    ymin=0.0,
    ymax=1.0,
    xmin=0.0,
    xmax=1.0,
    tick pos=left,
    domain=0:1,
    legend pos={north west},
    legend style={draw=none, fill=none, font=\sffamily\small},
    legend cell align={left},
    legend image post style={xscale=0.5},
    style = {font=\scriptsize},
    xlabel style = {font=\small},
]
    \addplot[color=gray, densely dashed, line width=1.1pt]{x};
    \addlegendentry{threshold}
    \addplot[color=blue, line width=1.1pt] table [
    x = threshold,
    y = P-values-true,
    ]{\recursivedata};
     \addlegendentry{true}
    \addplot[color=orange, line width=1.1pt] table [
    x = threshold,
    y = P-values-est,
    ]{\recursivedata};
     \addlegendentry{estimate}
\end{axis}
\end{tikzpicture}
         \caption{$\Vp(s_0 \sep {\hat \pi})$ vs $\theta$}
\end{subfigure}
\begin{subfigure}{0.49\columnwidth}
\begin{tikzpicture}
%threshold,P-values-est,P-values-true,P-values-horizon-1,threshold,V-values-est,V-values-true
\begin{axis}[
    width=1.2\textwidth,
    xlabel={$\theta$},
    axis lines=left,
    xmin=0.0,
    xmax=1.0,
    ymin=-2.0,
    ymax=-1.4,
    tick pos=left,
    legend pos={north west},
    legend style={draw=none, fill=none, font=\sffamily\small},
    legend cell align={left},
    legend image post style={xscale=0.5},
    style = {font=\scriptsize},
    xlabel style = {font=\small}
]
    \addplot[color=blue, line width=1.1pt] table [
    x = threshold,
    y = V-values-true,
    ]{\recursivedata};
     \addlegendentry{true}
    \addplot[color=orange, line width=1.1pt] table [
    x = threshold,
    y = V-values-est,
    ]{\recursivedata};
     \addlegendentry{estimate}
\end{axis}
\end{tikzpicture}
        \caption{$\Vr(s_0 \sep {\hat \pi})$ vs $\theta$}
\end{subfigure}
%      \begin{subfigure}{0.49\textwidth}
%         \adjincludegraphics[clip, width=\textwidth]{Figures/Our_VI_P.png}
%         \caption{$\Vp(s_0 \sep {\hat \pi})$ vs $\theta$}
%      \end{subfigure}
%      \begin{subfigure}{0.49\textwidth}
%         \adjincludegraphics[clip, width=\textwidth]{Figures/Our_VI_Q.png}          
%         \caption{$\Vr(s_0 \sep {\hat \pi})$ vs $\theta$}
%      \end{subfigure}
      \caption{Experimental results of our proposed approach}
      \label{fig:OurVI}
\end{minipage}
\end{figure*}

Now that we have a well-defined solution $\pi^N$, the next question is ``how to find it?''. To be precise, the remaining issue is ``how to estimate all the action-value functions $\optQp{}^n$ ($n$-horizon) and $\optQr{}^n$ (infinite-horizon) for $n = 1, 2, 3, \dots, N$?''. The answer is that reinforcement learning ideas, such as value iteration and Q-learning, can be employed for such a purpose. In what follows, we compare naive value iteration based on the Bellman operator~$\mathcal{T}_\theta$, which does not hold the fixed point property $\mathbf{P4}$, with the idea of recursive constraints and bounded probabilistic reachability presented in this section.

\section{Experiments}\label{sec:experiments}

To illustrate the benefits of our approach, we implemented competing value iteration methods and applied them to a ``cliffworld'' environment~\cite{Sutton2018} that generalizes the MDP in~\Cref{fig:counter-example} by allowing actions in all possible directions; transitions to the desired direction are made with probability $0.5$ and to a random direction with the remaining $0.5$ probability. Using the full range of $\theta$, in~\Cref{fig:NaiveVI} we present the results of naive value iteration (\Cref{alg:naive VI}) with a total of 50 iterations. In \Cref{fig:OurVI} we present corresponding results using our proposed approach (\Cref{alg:recursive VI}) with a total of 15 iterations and horizon $N=15$. \Cref{fig:NaiveVI} shows that within a wide range around $\theta = 0.8$: (i) naive value iteration thinks that the current policy is safe (orange curve) while it actually is not (blue curve); (ii) the switching between the policies randomly induces errors, with the values not converging. In contrast, \Cref{fig:OurVI} shows that our approach does not exhibit any such chattering or violations. More experimental and theoretical study is ongoing work, but we can clearly see from the present experiments that the solution of our approach for each $\theta$ is apparently stable and converged.

%Idea: change (i) the iteration axis to horizon; (ii) the infinite horizon reachability with finite one; (iii) approximate the finite horizon reachability by using that at the previous horizon. 

% \section{Reinforcement Learning with Recursive Constraints}

% More investigation remains as a future work.

\balance

\section{Conclusion}\label{sec:conclusion}

In this work, we have considered a constrained optimization problem that arises naturally in the context of safety-critical systems.
Despite this, and our problem's apparently reasonable requirements, we find that there is no existing approach to adequately address it.
On closer inspection, we have discovered that our insistence on a deterministic policy and uniform optimality imposes restrictions that make an already hard task~\cite{Feinberg2000} more difficult.
In particular, we have shown with a simple example that a naive learning approach can be unstable and have no fixed point.
We have thus proposed and demonstrated a model-based reinforcement learning algorithm that uses recursive constraints to address the instability.
The approximative form of this approach reduces computational cost, while making the constraints conservative.

In future work, we will adapt our algorithm to the model-free case, and explore the possibility of using our approach with function approximation.

\iffalse
No one tells us the appropriate horizon. Long horizon increases computations.

Conservativm.... (see the experimental results first and then mention here if needed).

Mentions the future works: considering different properties (axioms) such as Pareto efficiency and, I call, maximal optimality (describing what to do when $\mathrm{argmax}$- or $\mathrm{argmin}$-set is not singleton), all together with P1--P4.

Develop a better algorithm that efficiently mitigate both conversatism and instability, potentially by bridging the gap between the properties P1 and P4.
\fi

%%%%%%%%%%%%%%%%%%%%%%%%%%%%%%%%%%%%%%%%%%%%%%%%%%%%%%%%%%%%%%%%%%%%%%%%

%%% The acknowledgments section is defined using the "acks" environment
%%% (rather than an unnumbered section). The use of this environment 
%%% ensures the proper identification of the section in the article 
%%% metadata as well as the consistent spelling of the heading.

% If you wish to include any acknowledgments in your paper (e.g., to 
% people or funding agencies), please do so using the `\texttt{acks}' 
% environment. Note that the text of your acknowledgments will be omitted
% if you compile your document with the `\texttt{anonymous}' option.

\begin{acks}
The authors gratefully acknowledge the financial support of Japanese Science and Technology agency (JST) ERATO project JPMJER1603: HASUO Metamathematics for Systems Design. The authors also express their gratitude to Abhinav Grover, for his participation in early discussions and implementation.
\end{acks}

%%%%%%%%%%%%%%%%%%%%%%%%%%%%%%%%%%%%%%%%%%%%%%%%%%%%%%%%%%%%%%%%%%%%%%%%

%%% The next two lines define, first, the bibliography style to be 
%%% applied, and, second, the bibliography file to be used.

\bibliographystyle{ACM-Reference-Format} 
\bibliography{main}

%%%%%%%%%%%%%%%%%%%%%%%%%%%%%%%%%%%%%%%%%%%%%%%%%%%%%%%%%%%%%%%%%%%%%%%%
\newpage
\nobalance
\newcommand{\Aconst}{\mathscr{\hat A}}
\newcommand{\nextpolicy}{\pi}

\begin{algorithm}[!ht]
 \begin{spacing}{1.1}
  \KwIn{\\
\begin{minipage}[t]{0.5\columnwidth}
$\mathscr{M}:$ MDP~$(\mathscr{S^+}, \mathscr{A^+}, \transfn, \gamma, \mathscr{R})$\\
$\mathscr{F}_\perp:$ set of all failure states\\
$\theta:$ constraint threshold $\in [0, 1)$
\end{minipage}
\begin{minipage}[t]{0.35\columnwidth}
$k:$ number of iterations
\end{minipage}
\vspace{4pt}
}
  \KwOut{\\
  	$\nextpolicy:$ solution policy\\
  	$\Qr, \Qp:$ estimates of action-value functions $\Qr(\cdot, \nextpolicy)$ and $\Qp(\cdot, \nextpolicy)$\\
  	}
  \tcc{initialization}
%  \ForEach{$s a \in \mathscr{S} \times \mathscr{A}(s)$}{
    $\Qr(s, a)\gets \text{initial value, e.g. } 0\quad\forall(s,a)\in \mathscr{S} \times \mathscr{A}(s)$\\
    % $\Qr(s, a)\gets v\in[0,1]\quad\forall(s,a)\in \mathscr{S} \times \mathscr{A}(s)$\\
$\Qp(s, a)\gets \text{initial value}\in[0, 1] \quad\forall(s,a)\in\mathscr{S} \times \mathscr{A}(s)$\\
    % $\Qp(s, a)\gets v\in[0, 1] \quad\forall(s,a)\in\mathscr{S} \times \mathscr{A}(s)$\\
%  }
%  \tcc{for each terminal state-actions}
%  \ForEach{ $sa \in \mathscr{S_\perp} \times \mathscr{A}(s)$}{
  	$\Qr(s, a) \gets \mathscr{R}(s, a, s)\quad\forall(s,a) \in \mathscr{S_\perp} \times \mathscr{A}(s)$\\
  	$\Qp(s, a) \gets \indicator(s \in \mathscr{F}_\perp)\quad\forall(s,a) \in \mathscr{S_\perp} \times \mathscr{A}(s)$\\
 	%$\Aconst(s) \gets \{ a \in \mathscr{A}(s) \mid \Qp(s, a) \leq \theta \} \quad \forall s \in \mathscr{S^+}$\\
	%$\nextpolicy \gets \text{GetPolicy}(\mathscr{S^+},  \Aconst,  \Qr,  \Qp)$\\
  \tcc{main loop}
  \DontPrintSemicolon
  \RepTimes{$k$}{
  %$\Delta \gets 0$\\
  $\Aconst(s) \gets \{ a \in \mathscr{A}(s) \mid \Qp(s, a) \leq \theta \}\quad \forall s \in \mathscr{S^+}$\\
  $\nextpolicy \gets\text{GetPolicy}(\Aconst,  \Qr,  \Qp)$\\
  \ForEach{ $(s,a) \in \mathscr{S} \times \mathscr{A}(s)$}{
  	$\displaystyle \optQr{}(s, a) \gets \sum_{s' \in \mathscr{S}^+} \mathscr{T}(s, a)(s') \cdot \big ( \mathscr{R}(s, a, s') + \gamma \cdot \Qr(s', \pi(s')) \big )\hspace{-1em}$\\
  	$\displaystyle \optQp{}(s, a) \gets \sum_{s' \in \mathscr{S}^+} \mathscr{T}(s, a)(s') \cdot \Qp(s', \pi(s')) $\\
  	%$\Delta \gets \max \left( \Delta\,, \big|\Qr(s, a) - \optQr{}(s, a)\big|, \big|\Qp(s, a) - \optQp{}(s, a) \big| \right)$
  }
    $\Qr \gets \optQr{}$\\
    $\Qp \gets \optQp{}$\\
}
\tcc{update the solution policy}
$\Aconst(s) \gets \{ a \in \mathscr{A}(s)\mid \Qp(s, a) \leq \theta \}\quad\forall s \in \mathscr{S^+}$\\
$\pi \gets \text{GetPolicy}(\Aconst, \Qr, \Qp)$\\
\Return $\nextpolicy, \Qr, \Qp$
\caption{Naive Value Iteration \label{alg:naive VI}}
\end{spacing}
\end{algorithm}

\begin{algorithm}[h]
\renewcommand{\algorithmcfname}{Subroutine}
\caption*{GetPolicy$(\Aconst,  Q,  \mathscr{P})$}\label{alg:update-policy}
\ForEach{$s\in\mathscr{S^+}$}{
%$\Aconst(s) \gets \{a' \in \mathscr{A}(s) \mid \Qp(s, a') \leq \theta \}$\\
$\nextpolicy(s)\gets a\in
\begin{cases}
        \displaystyle 
        \Argmax_{a' \in \Aconst(s)} Q(s, a') \;\, \text{ if } \Aconst(s) \neq \varnothing
        \\
        \displaystyle
        \Argmin_{a' \in \mathscr{A}(s)}\; \mathscr{P}(s, a') \;\, \text{ otherwise}
    \end{cases}$
}
\Return \nextpolicy
\end{algorithm}

\newpage

\begin{algorithm}[!h]
\begin{spacing}{1.1}
  \KwIn{\\
\begin{minipage}[t]{0.5\columnwidth}
$\mathscr{M}:$ MDP~$(\mathscr{S^+}, \mathscr{A^+}, \transfn, \gamma, \mathscr{R})$\\
$\mathscr{F}_\perp:$ set of all failure states\\
$\theta:$ constraint  threshold $\in [0, 1)$
\end{minipage}
\begin{minipage}[t]{0.35\columnwidth}
$k:$ number of iterations\\
$N:$ horizon $\in \mathbb{N}$\\
\end{minipage}
\vspace{4pt}
}
  \KwOut{\\
  	$\nextpolicy:$ solution policy\\
  	$\Qr^N, \Qp^N:$ estimates of functions $\Qr(\cdot, \nextpolicy)$ and $\Qp^N(\cdot, \nextpolicy)$\\
%   	$\Aconst^{1:N}(s):$ sequence of sets of available actions for all states $s$
  	}
  \tcc{initialization}
  	 $\Qr^{1:N}(s, a)\gets\text{initial value, e.g. } 0\quad\forall (s,a)\in \mathscr{S}\times\mathscr{A}(s)$\\
	 $\Qp^1(s, a) \gets \sum_{s' \in \mathscr{F}_\perp} \mathscr{T}(s, a)(s')\quad\forall (s,a)\in \mathscr{S}\times\mathscr{A}(s)$ \\
  	 $\Qr^{1:N}(s, a) \gets \mathscr{R}(s, a, s)\quad\forall (s, a) \in \mathscr{S_\perp} \times \mathscr{A}(s)$\\
	 $\Qp^{1:N}(s, a) \gets \indicator(s \in \mathscr{F}_\perp)\quad\forall (s, a) \in \mathscr{S_\perp} \times \mathscr{A}(s)$\\  	   	 

  \tcc{main loop}
  \DontPrintSemicolon
  \RepTimes{$k$}{
  %$\Delta \gets 0$\\
  $\Aconst \gets \mathscr{A}$\\
  \For{$n=1,2,\dots,N$}{
    $\Aconst(s) \gets \{ a \in \Aconst(s)\mid \Qp^n(s, a) \leq \theta \}\quad\forall s \in \mathscr{S^+}$\\
    $\nextpolicy \gets\text{GetPolicy}(\Aconst, \Qr^n, \Qp^n)$\\
  	    \ForEach{ $(s,a) \in \mathscr{S} \times \mathscr{A}(s)$}{
  	$\displaystyle \optQr{}(s, a) \gets \!\!\sum_{s' \in \mathscr{S}^+}\! \mathscr{T}(s, a)(s') \cdot \big ( \mathscr{R}(s, a, s') + \gamma \cdot \Qr^n(s', \pi(s')) \big )\hspace{-4em}$\\
  	%$\Delta \gets \max \left( \Delta\,, \;\left |\Qr^n(s, a) - \optQr{}(s, a)\right| \right)$
  	}
  	%$\Qr^n \gets \optQr{}$\\
	  \ForEach{ $(s, a) \in \mathscr{S} \times \mathscr{A}(s) \textbf{~\emph{if}~} n < N$}{
    	${\displaystyle \Qp^{n+1}(s, a) \gets \!\!\sum_{s' \in \mathscr{S}^+} \!\mathscr{T}(s, a)(s') \cdot \Qp^n(s', \pi(s'))}$ \\
	}
	$\Qr^n \gets \optQr{}$\\
  }
}
\tcc{update the solution policy}
$\Aconst \gets \mathscr{A}$\\
  \For{$n=1,2,\dots,N$}{
    $\Aconst(s) \gets \{ a \in \Aconst(s)\mid \Qp^n(s, a) \leq \theta \}\quad\forall s \in \mathscr{S^+}$\\
}
%$\Aconst(s) \gets \{ a \in \mathscr{A}(s)\mid \Qp^n(s, a) \leq \theta \}\quad\forall (s,n) \in \mathscr{S^+}\times\{1,\dots,N\}\hspace{-1em}$\\
$\pi \gets \text{GetPolicy}(\Aconst, \Qr^N, \Qp^N)$\\
\Return $\nextpolicy, \Qr^N, \Qp^N$
\caption{Value Iteration with Recursive Constraints \label{alg:recursive VI}}
\end{spacing}
\end{algorithm}

\end{document}